\documentclass[a4paper]{article}

\usepackage{interspeech2013,amssymb,amsmath,stmaryrd,epsfig}
\usepackage{color}
\usepackage[bookmarks=true, bookmarksnumbered=true, bookmarksopen=true,
  unicode=true, colorlinks=true,
  linkcolor=blue,citecolor=blue,filecolor=blue,urlcolor=blue,
  pdfstartview=FitH
]{hyperref}

\DeclareMathOperator*{\argmax}{arg\,max}

\ninept

\def\reg{{\rm\ooalign{\hfil
     \raise.07ex\hbox{\scriptsize R}\hfil\crcr\mathhexbox20D}}}

\title{Multiple topic identification in telephone conversations}

\makeatletter
\def\name#1{\gdef\@name{#1\\}}
\makeatother
\name{{\em Xavier Bost$^1$, Marc El-Beze$^1$, and Renato De Mori$^{1,2}$}}

\address{$^1$LIA, University of Avignon, France \\
  $^2$McGill University, School of Computer Science, Montreal, Quebec,
  Canada \\
  {\small \tt \{xavier.bost, marc.elbeze, renato.demori\}@univ-avignon.fr}}

\begin{document}
\maketitle

\begin{abstract}
This paper deals with the automatic analysis of conversations between
a customer and an agent in a call centre of a customer care
service. The purpose of the analysis is to hypothesize themes about
problems and complaints discussed in the conversation. Themes are
defined by the application documentation topics. A conversation
may contain mentions that are irrelevant for the application purpose
and multiple themes whose mentions may be interleaved portions of a
conversation that cannot be well defined. Two methods are proposed for
multiple theme hypothesization. One of them is based on a cosine
similarity measure using a bag of features extracted from the entire
conversation. The other method introduces the concept of thematic
density distributed around specific word positions in a
conversation. In addition to automatically selected words, word
bigrams with possible gaps between successive words are also
considered and selected. Experimental results show that the results
obtained with the proposed methods outperform the results obtained
with support vector machines on the same data. Furthermore, using the
theme skeleton of a conversation from which thematic densities are
derived, it will be possible to extract components of an automatic
conversation report to be used for improving the service performance.
\end{abstract}
\noindent{\bf Index Terms}: multi-topic audio document classification,
human/human conversation analysis, speech analytics, distance bigrams

\vspace{2mm}

\noindent \textcolor{red}{\textbf{Cite as:}\\X.~Bost, M.~El-B\`eze, R. De Mori.\\\href{https://www.isca-speech.org/archive/interspeech_2013/i13_3756.html}{Multiple topic identification in telephone conversations.}\\14th Annual Conference of the International Speech Communication Association (INTERSPEECH 2013).}

\section{Introduction}

There has been a growing interest in recent years on speech technology
capabilities for monitoring telephone services.  In order to provide a
high level of efficiency and user satisfaction, there is a consensus
on the importance of improving systems for analysing human/human
conversations to obtain reports on customer problems and the way an
agent has solved them. With these reports, useful statistics can be
obtained on problem types and facts, the efficiency of problem
solutions, user attitude and satisfaction.

The application considered in this paper deals with the automatic
analysis of dialogues between a call centre agent who can solve
problems defined by the application domain documentation and a
customer whose behaviour is unpredictable. The customer is expected to
seek information and/or formulate complaints about the Paris
transportation system and services.

The application documentation is a concise description of problem
themes and basic facts of each theme considered as a conversation
topic. The most important speech analytics for the application are the
themes even if facts and related complementary information are also
useful.

A conversation may contain more than one semantically related
themes. Depending on the type of relation, some themes discussed in a
conversation may be irrelevant for the application task. For example,
a customer may inquiry about an object lost on a transportation mean
that was late. In such a case, the loss is a much more relevant theme
than the traffic state. A component of the automatic analysis is an
Automatic Speech Recognition (\textsc{asr}) system that makes
recognition errors. In the above example, failure to detect the
mention to traffic state can be tolerated if the mention to the loss
fact is correctly hypothesized but in general, all the mentioned
themes must be taken into account. This paper focuses on the detection
of relevant themes in a service dialogue. Conversations to be analysed
may be about one or more domain themes. Detecting the possibility of
having multiple themes and identifying them is important for
estimating proportions of customer problems involving related
topics. Different themes can be mentioned in disjoint discourse
segments. In some cases, mentions of different themes may coexist in
short segments or even in a single sentence. Even when different
themes are mentioned in non-overlapping discourse segments, segment
boundaries may be difficult to estimate because of the errors
introduced by the \textsc{asr} system and the imprecise knowledge
about the language structures used by casual users in the considered
real-world situation. In spite of the just discussed difficulties, it
is worth considering the possibility of extracting suitable features
for theme detection considering that it is likely to have several
mentions of a theme in a conversation.

The paper is structured as follows. Section~\ref{sec:rel_work} of the
paper discusses related work.  Section~\ref{sec:app_feat} introduces
the application domain and the features used for theme
hypothesization. The first of two approaches to multiple theme
identification is introduced in section~\ref{sec:cosine}. It is based
on the automatic estimation of decision parameters applied to global
cosine similarity measures. The second approach, described in
section~\ref{sec:dens}, introduces the concept of dialogue skeleton
based on which thematic densities of features can be computed and used
as soft detections of possibly overlapping locations in a dialogue
where one or more themes are mentioned. Decision making with this
approach is also described. Experimental results are presented in
Section~\ref{sec:exp}.

\section{Related work}

\label{sec:rel_work}

Human/human spoken conversation analyses have been recently reviewed
in~\cite{ES1}. Methods for topic identification in audio documents are
reviewed in~\cite{ES2}. Solutions for the detection of conversation
segments expressing different topics have been proposed in many
publications, recently reviewed in~\cite{ES3}. Interesting solutions
have been proposed for linear models of non-hierarchical
segmentation. Some approaches propose inference methods for selecting
segmentation points at the local maxima of cohesion functions. Some
functions use features extracted in each conversation sentence or in a
window including a few sentences. Some search methods detect cohesion
in a conversation using language models and some others consider
hidden topics shared across documents.

A critical review on lexical cohesion can be found in~\cite{ES4} who
propose an unsupervised approach for hypothesizing segmentation points
using cue phrases automatically extracted from unlabelled data.  An
evaluation of coarse-grain segmentation can be found in~\cite{ES5}.

Multi-label classification is discussed in~\cite{ES6} and~\cite{ES7}
mostly for large collections of text documents. Particularly
interesting is a technique called creation consisting in creating new
composite labels for each association of multiple labels assigned to
an instance.

This paper extends concepts found in the recent literature by
introducing a version of bigram words that may have a gap of one
word. These features are used in decision strategies based on a
constrained application of the cosine similarity and on a new
definition of soft theme density location in a conversation.

\section{Application task and features}

\label{sec:app_feat}

The application task is about a customer care service
(\textsc{ccs}). The purpose of the application is to monitor the
effectiveness of a call centre by evaluating proportions of problem
items and solutions. Application relevant information is described in
the application requirements focussing on an informal description of
useful speech analytics not completely defined. The requirements and
other application documentation essentially describe the types of
problems a \textsc{ccs} agent is entitled to solve.

This paper proposes a new approach for automatically annotating
dialogues between an agent and a customer with one or more application
theme labels belonging to the set $\mathbb{C}$ defined as follows:

$\mathbb{C} :=$ \{~itinerary, lost and found, time schedules,
transportation card, traffic state, fine, special offers~\}.

Given a pair $(d, c)$, where $d \in \mathbb{X}$, a spoken dialogue in
the corpus $\mathbb{X}$, is described by a vector $v_d$ of features
and $c \in \mathbb{C}$ is a class corresponding to a theme described
by a vector $v_c$ of features of the same type as $v_d$, two
classification methods, indicated as $\gamma_1$ and $\gamma_2$, are
proposed for multiple theme classification.

With the purpose of increasing the performance of automatic multiple
theme hypothesization, bigrams were added to the lexicon of 7217 words
with the possibility of having also distance bigrams made of pairs of
words distant a maximum of two words. The feature set increases to
160433 features with this addition. In order to avoid data sparseness
effects, a reduced feature set $V$ was obtained by selecting features
based on their purity and coverage. Purity of a feature $t$ is defined
with the Gini criterion as follows:

\begin{equation}
    G(t) = \sum_{c \in \mathbb{C}} \mathbb{P}^2(c|t)
    = \sum_{c \in \mathbb{C}} \left ( \frac{df_c(t)}{df_{\mathbb{T}}(t)} \right ) ^2
\end{equation}

where $df_{\mathbb{T}}(t)$ is the number of dialogues of the train set
$\mathbb{T}$ containing term $t$ and $df_c(t)$ is the number of
dialogues of the train set containing term $t$ in dialogues annotated
with theme $c$.

A score $w_c(t)$ is introduced for feature $t$ in the entire train set
collection of dialogues discussing theme $c$. It is computed as
follows:

\begin{equation}
  w_c(t) = df_c(t) . idf^2(t) . G^2(t)
\end{equation}

where $idf(t)$ is the inverse document frequency for feature~$t$.

\section{Using a global cosine similarity measure}

\label{sec:cosine}

The classical cosine measure of similarity between the two vectors
$v_d \ (d \in \mathbb{X})$ and $v_c \ (c \in \mathbb{C})$ is defined
as:

\begin{equation}
    \text{sc}(d, c) = \cos(\widehat{v_d,v_c})
    = \frac{\sum_{t \in d \cap c}w_d(t).w_{c}(t)}{\sqrt{\sum_t w_d(t)^2.w_{c}(t)^2}}
\end{equation}

where $w_d(t)$ is a score for feature $t$ in dialogue
$d$.

Let $\gamma_1(d)$ be the set of themes discussed in dialogue $d$. A
first decision rule for automatically annotating dialogue $d$ with a
theme class label $c$ is:

\begin{equation}
  c \in \gamma_1(d) \Longrightarrow \text{sc}(d, c) \geqslant \rho.
  \text{sc}(d, \hat{c})
\end{equation}

where $\hat{c} := \argmax_{c' \in \mathbb{C}}\text{sc}(d, c')$; and $\rho
\in [0; 1]$ is an empirical parameter whose value is estimated by
experiments on the development set.

If the score of $\hat{c}$ is too low, then the application of the
above rule is not reliable. To overcome this problem, the following
additional rule is introduced:

\begin{equation}
  c \in \gamma_1(d) \Longrightarrow \text{sc}(d, c) \geqslant v
  \sum_{c' \in \mathbb{C}} \text{sc}(d, c')
\end{equation}

where $v \in [0; 1]$ is another parameter whose value is estimated by
experiments on the development set.

\subsection{Parameter estimation}

The values of parameters $\rho$ and $v$ have been estimated using 20
subsets of 98 dialogues each belonging to the development set and
selected with the same proportion of single and multiple theme
dialogues as in the development set. In order to estimate the optimal
values $\hat{\rho}$ and $\hat{v}$ of these two parameters, the
following decision rule has been applied:

\begin{equation}
  (\hat{\rho}, \hat{v}) = \argmax_{(\rho, v) \in [0; 1]^2} \sum_{i=1}^{20}
  F(\gamma_{1 (\rho, v)}, \mathbb{D}_i)
\end{equation}

where $F(\gamma_{1(\rho,v)}, \mathbb{D}_i)$ is the F-score (defined
in the following subsection) computed using model $\gamma_1$ with the
estimated values of $\rho$ and $v$ in the subset $\mathbb{D}_i \subset
\mathbb{D}$.

The optimal value of $\rho$, the proportion of the highest score
required for assigning themes to a dialogue, has been evaluated to
$\hat{\rho} = 0.69$, and the optimal value of $v$, the threshold
required, to $\hat{v} = 0.16$.

\subsection{Performance measures}

\label{subsec:metrics}

The proposed approaches have been evaluated following procedures
discussed in~\cite{ES7} with measures used in Information Retrieval
(\textsc{ir}) and accuracy as defined in the following for a corpus
$\mathbb{X}$ and a decision strategy $\gamma$.

Of particular importance is the F-score: based on this measure, and as
mentioned in subsection~\ref{subsec:res_an}, it is possible to find
the best trade-off between precision and recall by rejecting some of
the dialogues.

\bigskip

Recall~:

\begin{equation}
  R(\gamma, \mathbb{X}) = \frac{1}{|\mathbb{X}|} \sum_{d \in \mathbb{X}}
  \frac{|\gamma(d) \cap L(d)|}{|L(d)|}
\end{equation}

where $L(d)$ indicates the set of theme labels annotated for
conversation $d$.

Precision~:

\begin{equation}
  P(\gamma, \mathbb{X}) = \frac{1}{|\mathbb{X}|} \sum_{d \in \mathbb{X}}
  \frac{|\gamma(d) \cap L(d)|}{|\gamma(d)|}
\end{equation}

F-score~:

\begin{equation}
  F(\gamma, \mathbb{X}) = \frac{2P(\gamma, \mathbb{X})R(\gamma,
    \mathbb{X})}{P(\gamma, \mathbb{X})+R(\gamma, \mathbb{X})}
\end{equation}

Accuracy~:

\begin{equation}
  A(\gamma, \mathbb{X}) = \frac{1}{|\mathbb{X}|} \sum_{d \in \mathbb{X}}
  \frac{|\gamma(d) \cap L(d)|}{|\gamma(d) \cup L(d)|}
\end{equation}

\section{Automatic annotation based on thematic densities}

\label{sec:dens}

\subsection{Thematic density}

The contribution to theme $c$ of the features at the
$i$-th location in a dialogue is:

\begin{equation}
w_c(p_i) = \frac{1}{\parallel v_c \parallel} \sum_{t \in T_{p_i}}
w_c(t) \ \  (i = 1, ..., n)
\end{equation}

where $T_{p_i}$ is the set made of the $i$-th word in a conversation
and the bigrams associated with it.

A thematic density $d_c(p_i)$ of theme $c$ is associated with position
$i$ and is defined as follows:

\begin{equation}
d_c(p_i) = \frac{\sum_{j=1}^n
  \frac{w_c(p_j)}{\lambda^{d_j}}}{\sum_{j=1}^n
  \frac{1}{\lambda^{d_j}}} \ (i = 1, ..., n)
\end{equation}

where $\lambda \geqslant 1$ is a parameter of sensitivity to
proximity whose value is estimated by experiments on the
  development set and $d_j := |i - j|$.

\subsection{Dialogue skeleton}

The theme density at a specific dialogue location makes it possible to
derive a thematic skeleton of a dialogue.

\begin{figure}[t]
  \centering
  \includegraphics[width=90mm]{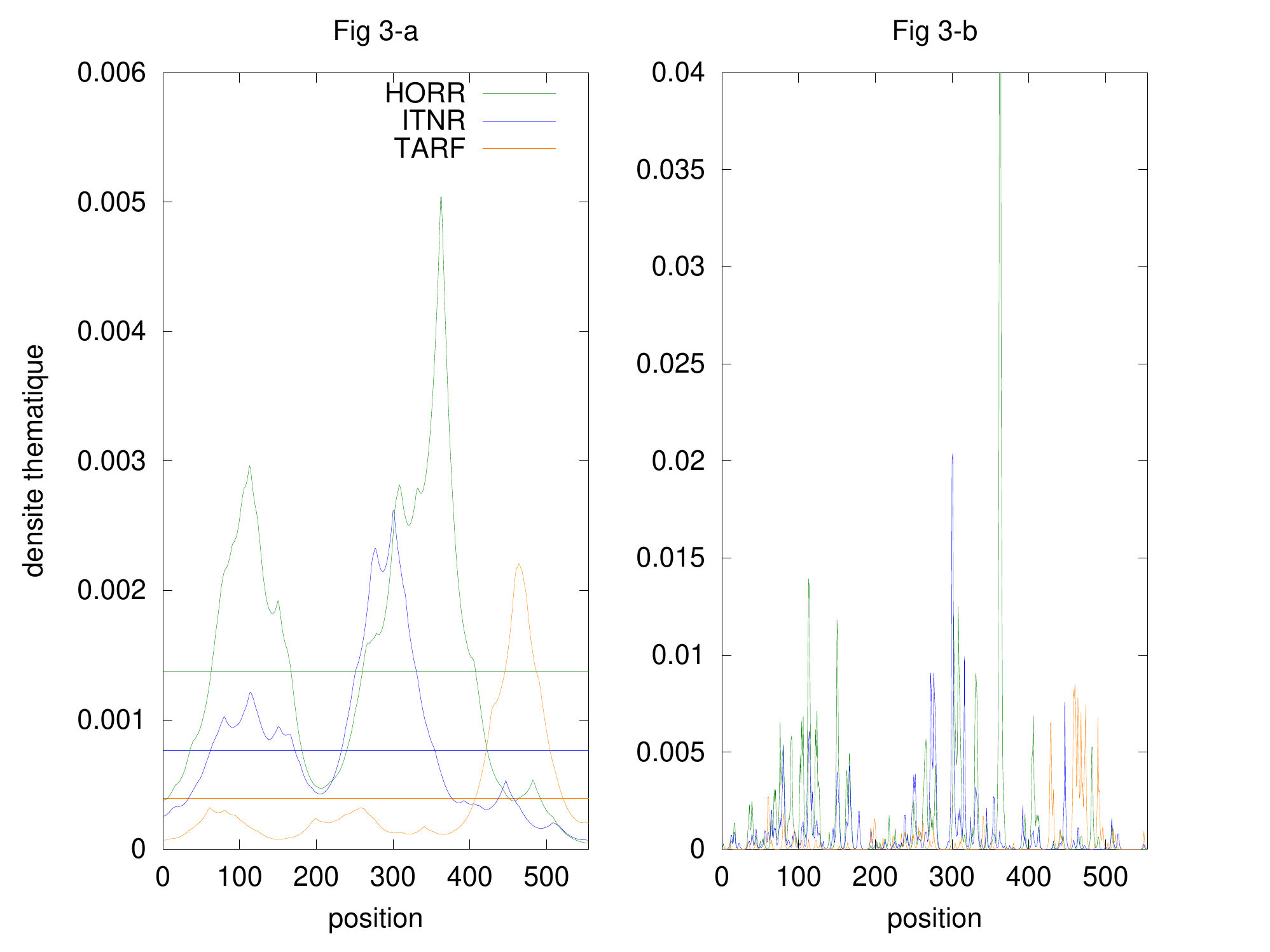}
  \caption{{\it Thematic densities as function of location in a dialogue
      skeleton. Densities are plotted for $\lambda = 1$ and $\lambda =
      1.05$ (fig 1- a) and $\lambda = 2.8$ (fig 1- b) and three themes:
      schedule (indicated as \textsc{horr}), fare (indicated as
      \textsc{tarf}) and itinerary (indicated as \textsc{itnr}.}}
  \label{lambda}
\end{figure}

Figure~\ref{lambda} shows the skeleton of a dialogue obtained with an
automatic transcription for $\lambda = 1$ and $\lambda = 1.05$ (fig 1-
a) and $\lambda = 2.8$ (fig 1- b). The figures plot the thematic
density as function of the dialogue location measured in numbers of
words preceding the location. The conversation is about a request of
bus \textit{schedule} (indicated as \textsc{horr}) and a \textit{fare}
(indicated as \textsc{tarf}).  Three functions are plotted. Two for
these themes and a third for the theme \textit{itinerary} (indicated
as \textsc{itnr}).

As an example, an excerpt from the dialogue obtained from manual
transcriptions is reported in the following. The dialogue positions
corresponding to each turn are in brackets.

\begin{itemize}

\item[--] \textit{Customer} [82--100]~: I would like to know if busses
  start running in early morning.

\item[--] \textit{Agent} [101--118]~: First start is at 5~:45 at Opera
  square.

\item[--] (...)
  
\item[--] ~\textit{Agent} [304--314]~: It will be there at 8~:48.

\item[--] (...)

\item[--] \textit{Agent} [442--470]~: eh no .... There are specific
  fares for the airport. I'll give you the amount .... Fare is 9 euros
  10.

\item[--] \textit{Customer} [471--496]~: 9 euros 10, should we pay 9
  euros 10 cash.

\end{itemize}

With $\lambda = 2.8$ (fig~1-~b), distant contexts tend to be
neglected. In this case decision may be adversely affected by isolated
features that may not be relevant for theme hypothesization. For
example, the expression \textit{there} (``la-bas'' in French) in turn
[304--314] tends to show more evidence for itinerary with a peak of
density at location 300, while it is used here in a request of
schedule.

Local context is more appropriately taken into account with $\lambda =
1.05$ (fig~1-~a) with the result of reducing the relevance of the
itinerary hypothesis just supported by the word \textit{there}. When
$\lambda = 1$ (horizontal lines in fig~1-~a), close contexts tend to
be neglected giving more importance to global features used in the
approach introduced in Section~\ref{sec:cosine}.

In conclusion, with a well-suited value of $\lambda$ (=1.05), thematic
coherence tends to make decisions more accurate.

\subsection{Decision making}

A theme $c$ is considered as discussed in a dialogue $d$ if it has
dominant density at a location of the dialogue (rule
  \eqref{eq:RL1}) and if the sum of its densities in positions where
  it is dominant exceeds an empirically determined threshold (rule
  \eqref{eq:RL2})~:

\begin{equation}
  c \in \gamma_2(d) \Rightarrow \exists i \in \llbracket {1};{n}
  \rrbracket: \forall c' \in \mathbb{C}, \ d_c(p_i) \geqslant
  d_{c'}(p_i)
  \label{eq:RL1}
\end{equation}

\begin{equation}
  c \in \gamma_2(d) \Longrightarrow \sum_{i \in I} d_c(p_i) \geqslant
  v \sum_{i=1}^n d_{c_i}(p_i) 
  \label{eq:RL2}
\end{equation}

where $v \in [0, 1]$ is a parameter whose value is estimated by
experiments on the developement set; $c_i$ is the theme of dominant
density at the $i$-th position in the dialogue; and $I := \{n \in
\mathbb{N} \ | \ \forall c' \in \mathbb{C}, \ d_c(p_n) \geqslant
d_{c'}(p_n) \}$.

\section{Experiments}

\label{sec:exp}

\subsection{Experimental framework}

Experiments have been performed using an \textsc{asr} system described
in~\cite{ES8}. It is based on triphone acoustic hidden Markov models
(\textsc{hmm}) with mixtures of Gaussians belonging to a set of 230000
distributions. Model parameters were estimated with maximum \textit{a
  posteriori} probability (\textsc{map}) adaptation of 150 hours of
speech in telephone bandwidth with the data of the train set. A corpus
of 1658 telephone conversations was collected at the call centre of
the public transportation service in Paris. The corpus is split into a
train, a development and a test set containing respectively 884, 196
and 578 conversations. A 3-gram language model (\textsc{lm}) was
obtained by adapting with the transcriptions of the train set a basic
\textsc{lm}. An initial set of experiments were performed with this
system resulting with an overall \textsc{wer} on the test set of 57\%
(52\% for agents and 62\% for users). These high error rates are
mainly due to speech disfluencies and to adverse acoustic environments
for some dialogues when, for example, users are calling from train
stations or noisy streets with mobile phones. Furthermore, the signal
of some sentences is saturated or of low intensity due to the distance
between speakers and phones.

The annotation with possible multiple themes of the development and
test corpora has been performed in a batch process by maximizing the
agreement between three annotators.

It is important to notice that the training corpus dialogues have been
labelled on the fly by the agents with the constraint to choose one
and only one theme corresponding to the main customer concern. When
this was not clear, the annotation was based on the problem expressed
at the beginning of the conversation. With such a procedure it is not
possible to create bi-labels as described in~\cite{ES6} and~\cite{ES7}
and mentioned in Section~\ref{sec:rel_work}.

\subsection{Evaluation}

For the sake of comparison, the results obtained with the proposed
classification approaches have been compared with the results obtained
with a support vector machine (\textsc{svm}) using the same features
(unigrams and bigrams with possible gap of one word) and a linear
kernel. For every theme $c_i \in \mathbb{C}$, a binary classifier
$\gamma_i : \mathbb{X} \rightarrow \{c_i, \overline{c_i} \}$ is
defined. Every pair $(d, c_i) \in \mathbb{X} \times \mathbb{C}$ is
associated with the score computed by this classifier. The candidate
theme hypotheses for a conversation are those whose score is in an
interval corresponding to an empirically determined proportion of the
highest one. In addition to that, the hypothesis with the highest
score must be above a threshold empirically determined for this
purpose.

Results obtained with the cosine measure and with the theme density
are reported in Table~\ref{table_res_dev} for the development set and
in Table~\ref{table_res_test} for the test set. \textsc{man} and
\textsc{asr} respectively indicate manual transcriptions and automatic
transcriptions obtained with the most likely sequence of word
hypotheses generated by the \textsc{asr} system.

\begin{table} [th]
\caption{\label{table_res_dev} {\it Results obtained with the \textsc{svm}, the
    cosine measure and the theme density for the development set.}}
\vspace{2mm}
\centerline{
\begin{tabular}{|c||c|c|c||c|c|c||}
    \hline
    \textsc{dev} & \multicolumn{3}{|c||}{\textsc{man}} & \multicolumn{3}{|c||}{\textsc{asr}} \\
    \cline{2-7}
    & \textbf{\textsc{svm}} & \textbf{cos.} & \textbf{dens.} & \textbf{\textsc{svm}} & \textbf{cos.} & \textbf{dens.} \\
    \hline
    \textbf{Accuracy} & 0.77 & 0.85 & 0.85 & 0.70 & 0.80 & 0.81 \\
    \hline
    \textbf{Precision} & 0.88 & 0.92 & 0.94 & 0.79 & 0.87 & 0.90 \\
    \hline
    \textbf{Recall} & 0.85 & 0.92 & 0.88 & 0.81 & 0.89 & 0.86 \\
    \hline
    \textbf{F-score} & 0.86 & \textbf{0.92} & 0.91 & 0.80 & \textbf{0.88} & \textbf{0.88} \\
    \hline
\end{tabular}}
\end{table}

\begin{table} [th]
\caption{\label{table_res_test} {\it Results obtained with the
    \textsc{svm}, the cosine measure and the theme density for the
    test set. For the cosine-based method, the estimated confidence
    interval is $\pm$ 2.74\% for manual transcriptions, and $\pm$ 3\%
    for the \textsc{asr} output.}}
\vspace{2mm}
\centerline{
\begin{tabular}{|c||c|c|c||c|c|c||}
    \hline
    \textsc{test} & \multicolumn{3}{|c||}{\textsc{man}} & \multicolumn{3}{|c||}{\textsc{asr}} \\
    \cline{2-7}
    & \textbf{\textsc{svm}} & \textbf{cos.} & \textbf{dens.} & \textbf{\textsc{svm}} & \textbf{cos.} & \textbf{dens.} \\
    \hline
    \textbf{Accuracy} & 0.74 & 0.78 & 0.78 & 0.64 & 0.71 & 0.71 \\
    \hline
    \textbf{Precision} & 0.85 & 0.86 & 0.86 & 0.72 & 0.79 & 0.79 \\
    \hline
    \textbf{Recall} & 0.83 & 0.87 & 0.85 & 0.77 & 0.83 & 0.80 \\
    \hline
    \textbf{F-score} & 0.84 & \textbf{0.87} & 0.85 & 0.75 & \textbf{0.81} & 0.80 \\
    \hline
\end{tabular}}
\end{table}

\subsection{Results analysis}

\label{subsec:res_an}

The development set was collected in the same time period (fall) as
the train set, while the test set was collected in the summer. The
difference in the results obtained with the test and the development
sets can be explained in part considering the frequency of different
events in the two time periods (\textit{e.~g.} strikes in the fall and
specific maintenance works in the summer).

The strategies of all the used methods for hypothesizing a theme in
addition to the dominant one give better results on the hypothesized
latter theme compared to those obtained for the former one (monolabel
categorization). For example, using the relative value of the cosine
measure with the dev set, improvements from 0.88 to 0.92 using the
manual transcriptions and from 0.85 to 0.88 with the automatic
transcriptions are respectively observed for the F-score. The same
improvements are observed with the test set.

Using the development set for inferring a rejection rule based on
close scores between the first two hypotheses, an F-score of 0.83 is
obtained on the automatic transcriptions of the test set with a
rejection rate close to 10\%, the rate of disagreement between human
annotators.

\section{Conclusion and future work}

\label{sec:concl}

Features have been proposed for the hypothesization of one or more
themes mentioned in a conversation between a call centre agent and a
calling customer. They are sets of words, bigrams and distant bigrams
automatically selected in the application domain.

Two approaches have been proposed for multiple theme
hypothesization. A first approach is based on a cosine similarity
measure applied to the features extracted from a conversation and the
other based on a new definition of theme density obtained considering
a conversation skeleton. The approaches have been evaluated and have
shown to outperform an \textsc{svm} classifier using the same features
and data.

Future research will include the search for confidence measures
suitable for the multi topic task and the use of conversation
skeletons for extracting short reports on agent/customer
dialogues. From these reports, proportions of speech analytics will be
extracted and used for monitoring frequencies, importance and solution
rates for different type of facts and problems.

\section{Acknowledgements}

This work is supported by the French National Research Agency
(\textsc{anr}), Project \textsc{decoda}, and the French business
clusters Cap Digital and \textsc{scs}. The corpus has been provided by
the \textsc{ratp} (Paris public transport company).

\newpage

\eightpt
\bibliographystyle{IEEEtran}

\end{document}